\documentclass[conference]{IEEEtran}
\IEEEoverridecommandlockouts

\usepackage{cite}
\usepackage{amsmath,amssymb,amsfonts}
\usepackage{algorithmic}
\usepackage{graphicx}
\usepackage{textcomp}
\usepackage[table,xcdraw]{xcolor}
\def\BibTeX{{\rm B\kern-.05em{\sc i\kern-.025em b}\kern-.08em
    T\kern-.1667em\lower.7ex\hbox{E}\kern-.125emX}}
\usepackage[]{todonotes}
\usepackage[hidelinks]{hyperref}
\usepackage{subfigure}
\usepackage{multicol}
\usepackage{tabularx}

\newcommand{\uu}{\text {uu}}

\newcommand{\uuss}{\frac{\text{uu}}{\text{s}^2}}
\newcommand{\rs}{\frac{\text{rad}}{\text{s}}}
\newcommand{\rss}{\frac{\text{rad}}{\text{s}^2}}

\begin{document}

\title{On the Verge of Solving Rocket League using Deep Reinforcement Learning and Sim-to-sim Transfer}

\author{
\IEEEauthorblockN{Marco Pleines\footnote{Correspondence to marco.pleines@tu-dortmund.de}, Konstantin Ramthun, Yannik Wegener, Hendrik Meyer, Matthias Pallasch, Sebastian Prior\\
Jannik Dr{\"o}gem{\"u}ller, Leon B{\"u}ttinghaus, Thilo R{\"o}themeyer, Alexander Kaschwig,\\
Oliver Chmurzynski, Frederik Rohkr{\"a}hmer, Roman Kalkreuth, Frank Zimmer\IEEEauthorrefmark{1}, Mike Preuss\IEEEauthorrefmark{2}}
\IEEEauthorblockA{
\textit{Department of Computer Science},
\textit{TU Dortmund University},
Dortmund, Germany \\
\IEEEauthorrefmark{1}\textit{Department of Communication and Environment, Rhine-Waal University of Applied Sciences}, Kamp-Linfort, Germany \\
\IEEEauthorrefmark{2}\textit{LIACS Universiteit Leiden}, Leiden, Netherlands \\
}
}


\maketitle

\begin{abstract}
Autonomously trained agents that are supposed to play video games reasonably well rely either on fast simulation speeds or heavy parallelization across thousands of machines running concurrently.
This work explores a third way that is established in robotics, namely sim-to-real transfer, or if the game is considered a simulation itself, sim-to-sim transfer.
In the case of Rocket League, we demonstrate that single behaviors of goalies and strikers can be successfully learned using Deep Reinforcement Learning in the simulation environment and transferred back to the original game.
Although the implemented training simulation is to some extent inaccurate, the goalkeeping agent saves nearly $100\%$ of its faced shots once transferred, while the striking agent scores in about $75\%$ of cases.
Therefore, the trained agent is robust enough and able to generalize to the target domain of Rocket League.
\end{abstract}

\section{Introduction}
\makeatletter
\newcommand*{\rom}[1]{\expandafter\@slowromancap\romannumeral #1@}
\makeatother
    
The spectacular successes of agents playing considerably difficult games, such as StarCraft~\rom{2} \cite{DBLP:journals/nature/VinyalsBCMDCCPE19} and DotA~2 \cite{DBLP:journals/corr/abs-1912-06680}, have been possible only because the employed algorithms were able to train on huge numbers of games on the order of billions or more.
Unfortunately, and despite many improvements achieved in AI in recent years, the utilized Deep Learning methods are still relatively sample inefficient. 
To deal with this problem, fast running environments or high amounts of computing resources are vital.
OpenAI~Five for DotA~2 \cite{DBLP:journals/corr/abs-1912-06680} is an example of the utilization of hundreds of thousands of computing cores in order to achieve high throughput in terms of played games.
However, this way is closed for games that run only on specific platforms and are thus very hard to parallelize.
Moreover, not many research groups have such resources at their disposal.
Video games that suffer from not being able to be sped up significantly, risk minimal running times and hence repeatability.
Therefore it makes sense to look for alternative ways to tackle difficult problems.

\IEEEpubid{\begin{minipage}{\textwidth}\ \\ \\[12pt] Accepted to IEEE CoG 2022 \end{minipage}}

\begin{figure}
\centering
\subfigure{\includegraphics[clip, trim=0 1cm 5cm 0, width=0.4\textwidth]{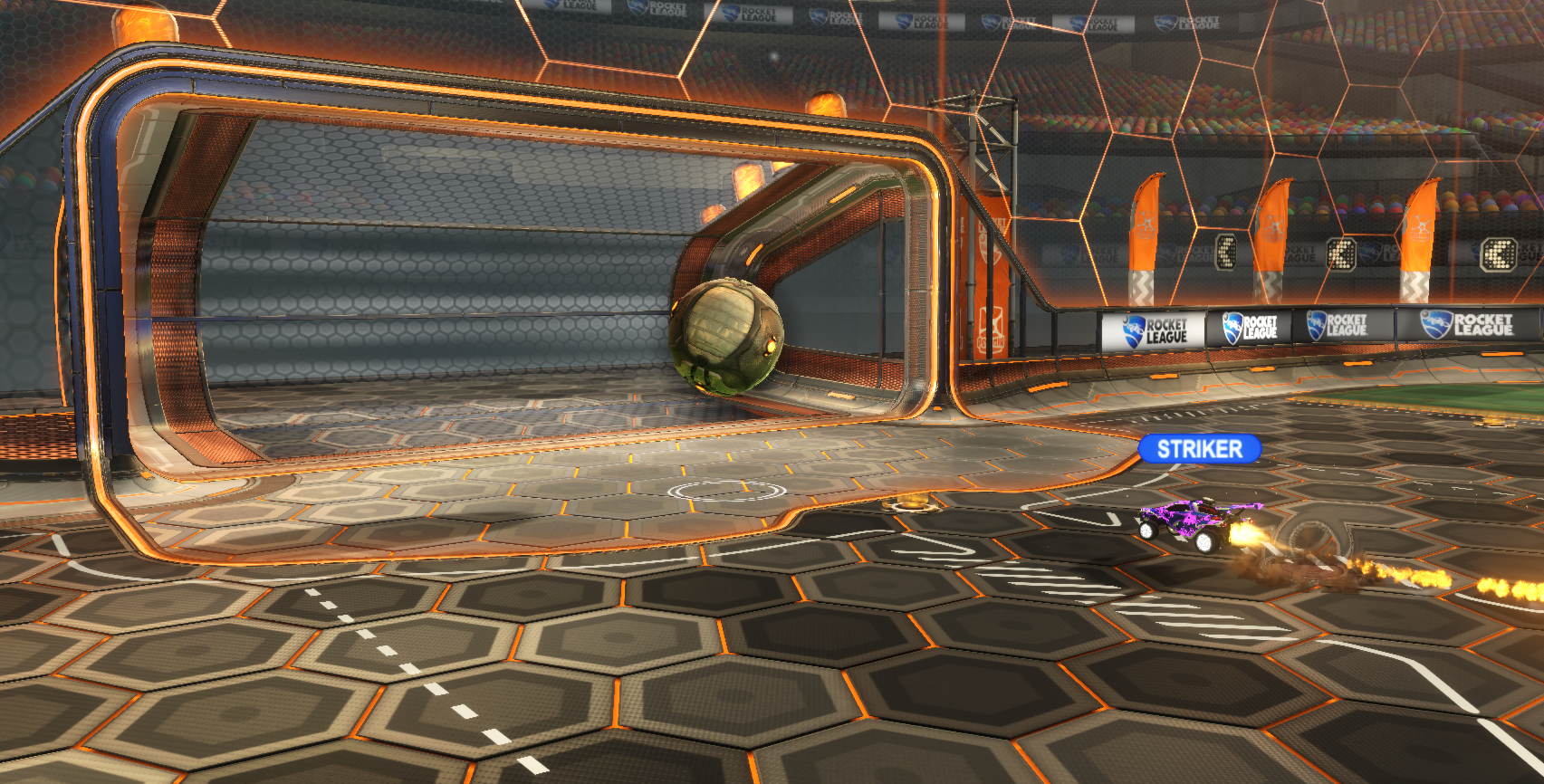}}
\\ \hspace{0.04cm}
\subfigure{\includegraphics[clip, trim=0 1cm 5.5cm 0, width=0.4\textwidth]{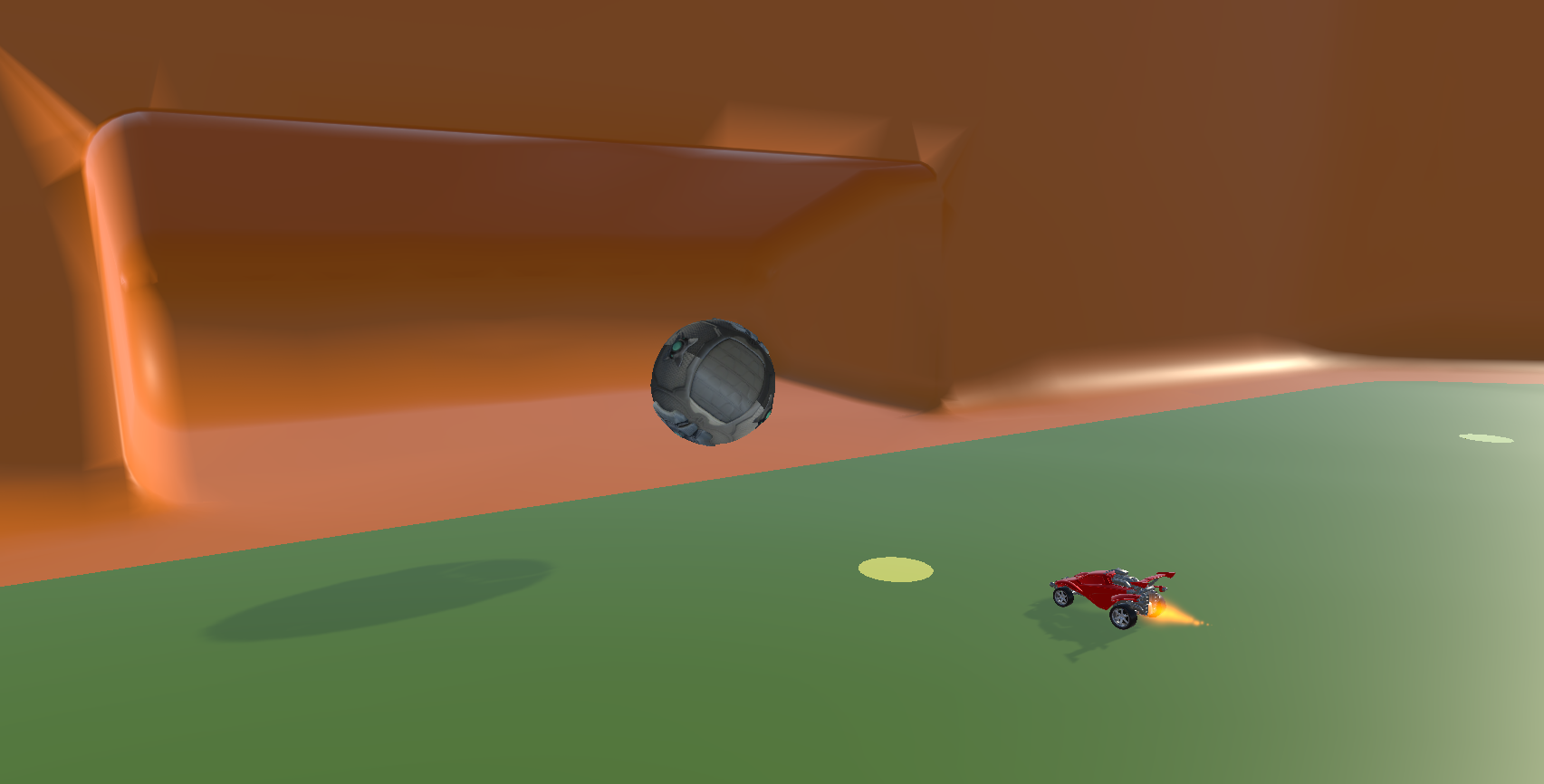}}
\caption{The game of Rocket League (top) and the contributed simulation (bottom), which notably advances its ancestor project \emph{RoboLeague} \cite{Roboleague2021}.}
\label{fig:both_sims}
\vspace{-0.15in}
\end{figure}

Sim-to-real transfer offers such an alternative way and is well established in robotics, and it follows the general idea that robot behavior can be learned in a very simplified simulation environment and the trained agents can then be successfully transferred to the original environment.
If the target platform is a game as well, we may speak of sim-to-sim transfer because the original game is also virtual, just computationally much more costly.
This approach is applicable to current games, even if they are not parallelizable, and makes them available for modern Deep Reinforcement Learning (DRL) methods.
There is of course a downside of this approach, namely that it may be difficult or even infeasible to establish a simulation that is similar enough to enable transfer later on, but still simple enough to speed up learning significantly.
A considerable amount of effort has to be invested in establishing this simulation environment before we can make any progress on the learning task. 

To our knowledge, the sim-to-sim approach has not yet been applied to train agents for a recent game. Therefore we aim to explore the possibilities of this direction in order to detect how simple the simulation can be, and how good the transfer to the original game works.

The game we choose as a test case of the sim-to-sim approach is Rocket League (Figure \ref{fig:both_sims}), which basically resembles indoor football with cars and teams of 3.
Rocket league is freely available for Windows and Mac, possesses a bot API (RLBot \cite{RLBotWiki2021}) and a community of bot developers next to a large human player base.
As the 3 members of each team control car avatars with physical properties different from human runners, the overall tactics are the one of rotation without fixed roles.
Thereby, large parts of the current speed can be conserved and players do not have to accelerate from zero when ball possession changes \cite{Verhoefen2020}.
Next to basic abilities attempting to shoot towards the goal and to move the goalie in order to prevent a goal, Rocket League is a minimal \emph{team AI} setting \cite{MozgovoyPB21} where layers of team tactics and strategy can be learned.


            

The first step of our work re-implements not all, but multiple physical gameplay mechanics of Rocket League using the game engine Unity, which results in a slightly inaccurate simulation.
We then train an agent in a relatively easy goalie and striker environment using Proximal Policy Optimization (PPO) \cite{Schulman2017}.
The learned behaviors are then transferred to Rocket League for evaluation.
Even though the training simulation is imperfect, the transferred behaviors are robust enough to succeed at their tasks by generalizing to the domain of Rocket League.
The goalkeeping agent saves nearly $100\%$ of the shots faced, while the striking agent scores about $75\%$ of its shots.
The sim-to-sim transfer is further examined by ablating physical adaptations that were added to the training simulation.

This paper proceeds with elaborating on related work.
Then, the physical gameplay mechanics of Rocket League are shown.
After illustrating the trained goalie and striker environment, PPO and algorithmic details are presented.
Section \ref{sec:sim-to-sim} examines the sim-to-sim transfer.
Before concluding our work, a discussion is provided.


\section{Related Work}

Sim-to-sim transfer on a popular multiplayer team video game touches majorly on two different areas, namely multi-agent and sim-to-real transfer.
DotA~2 and StarCraft~\rom{2} are the already mentioned prominent examples in the field of multi-agent environments.
As this work focuses on single-agent environments, namely the goalkeeper and striker environments, related work on sim-to-real transfer is focused next.

Given the real world, a considered prime example for multi-agent scenarios is \emph{RoboCup}.
RoboCup is an annual international competition~\cite{DBLP:conf/agents/KitanoAKNO97} that offers a publicly effective open challenge for the intersection of robotics and AI research.
The competition is known for the robot soccer cup but also includes other challenges.
Reinforcement Learning (RL) has been successfully applied to simulated robot soccer in the past~\cite{DBLP:journals/corr/HausknechtS15a} and has been found a powerful method for tackling robot soccer.
A recent survey \cite{AntonioniSRN21} provides insights into robot soccer and highlights significant trends, which briefly mention the transfer from simulation to the real world.

In general, sim-to-real transfer is a well-established method for robot learning and is widely used
in combination with RL.
It allows the transition of an RL agent's behavior, which has been trained in simulations, to real-world environments.
Sim-to-real transfer has been predominantly applied to RL-based robotics~\cite{DBLP:conf/ssci/ZhaoQW20} where the robotic agent has been trained with state-of-the-art RL techniques like PPO~\cite{Schulman2017}. 
Popular applications for sim-to-real transfer in robotics have been autonomous racing~\cite{DBLP:conf/icra/BalajiMGGDKRSTT20}, Robot Soccer~\cite{DBLP:journals/corr/abs-1911-01529},  navigation~\cite{DBLP:journals/corr/abs-1906-04452}, and control tasks~\cite{DBLP:conf/icra/PedersenMC20}. 
To address the inability to exactly match the real-world environment, a challenge commonly known as sim-to-real gap, steps have also been taken towards generalized sim-to-real transfer for robot learning~\cite{DBLP:conf/cvpr/RaoHILIK20,DBLP:conf/icra/HoRXJKB21}. 
The translation of synthetic images to realistic ones at the pixel level is employed by a method called GraspGAN~\cite{DBLP:conf/icra/BousmalisIWBKKD18} which utilizes a generative adversarial network (GAN)~\cite{DBLP:conf/nips/GoodfellowPMXWOCB14}. 
GANs are able to generate synthetic data with good generalization ability.
This property can be used for image synthesis to model the transformation between simulated and real images. GraspGAN provides a method called \textit{pixel-level domain adaptation}, which translates synthetic images to realistic ones at the pixel level.
The synthesized pseudo-real images correct the sim-to-real gap to some extent.
Overall, it has been found that the respective policies learned with simulations execute more successfully on real robots when GraspGAN is used~\cite{DBLP:conf/icra/BousmalisIWBKKD18}. 

Another approach to narrow the sim-to-real gap is domain randomization \cite{DomainRandomizationTobin2017}.
Its goal is to train the agent in plenty of randomized domains to generalize to the real domain.
By randomizing all physical properties and visual appearances during training in the simulation, a trained behavior was successfully transferred to the real world to solve the Rubik's cube \cite{RubikCube2019}.


\section{Rocket League Environment}

This section starts out by providing an overview of vital components of Rocket League's physical gameplay mechanics, which are implemented in the training simulation based on the game engine Unity and the ML-Agents Toolkit~\cite{Juliani2019}.
RLBot~\cite{RLBotWiki2021} provides the interface to Rocket League where the training situations can be reproduced.
Afterward, the DRL environments, designated for training, and their properties are detailed.
The code is open source\footnote{\url{https://github.com/PG642}}.

\subsection{Implementation of the Training Simulation}
\label{sec:env}

\renewcommand{\arraystretch}{1.2}
\begin{table*}[]
\caption{Overview on essential physical gameplay mechanics present in Rocket League, which are added to the training simulation.}
\begin{tabular}{lll}
\hline
\rowcolor[HTML]{EFEFEF}
\textbf{Physics Component}            & \textbf{Sources}                                                   & \textbf{Additional Information and Different Parameters} \\ \hline

Entity Measures (e.g. Arena)          & \cite{Roboleague2021, RLBotWiki2021}              & \begin{tabular}[c]{@{}l@{}}Car model Octane and its collision mesh is used\\ Radius of the ball is set to $93.15$\uu~(value in Rocket League $92.75$\uu)\end{tabular} \\ 

\rowcolor[HTML]{EFEFEF}
Car: Velocity, Acceleration, Boost    & \cite{Mish2019}                                   & No modifications done                                                                           \\
Car: Jumps, Double Jumps, Dodge Rolls & \cite{RocketScienceDodge2018, RLBotWiki2021}      & Raise max. angular velocity during dodge from $5.5\rs$ to $7.3\rs$\\
\rowcolor[HTML]{EFEFEF}
Car: Air Control                      & \cite{Mish2019}                                   & Adjust drag coefficients for roll to $-4.75$ and pitch to $-2.85$                                                                           \\
 
Bullet and Psyonix Impulse            & \cite{Mish2019, Cone2018} & \begin{tabular}[c]{@{}l@{}}Used for the ball-to-car interaction and car-to-car interaction. The impulse by the\\ bullet engine replaces the Unity one. Psyonix impulse is an additional impulse\\on the center of the ball, which allows a better prediction and control of collisions.\end{tabular}  \\

\rowcolor[HTML]{EFEFEF} 
Ball Bouncing                         & \cite{Mish2019}                                   & Within the bounce's computation a ball radius of $91.25$uu is considered.
\\

Friction (Air, Ground) and Drifting   & \cite{Cone2018}                      & A   drag of $-525\uuss$ is used, which is reduced by more than half when the car is upside down.                                                                                      \\
\rowcolor[HTML]{EFEFEF}
Ground Stabilization                  & \cite{RLBotWiki2021, Cone2018}                                   & The stabilization torque is denoted by an acceleration of $50\rss$.                                                                                                  \\

Wall Stabilization                    & \cite{RLBotWiki2021}                              & Raise sticky forces for wall stabilization to an acceleration of $500\uuss$                                                                                                  \\
\rowcolor[HTML]{EFEFEF}
Suspension                            & \cite{RocketScienceSuspension2018} \cite{Suspensions}                &  \begin{tabular}[c]{@{}l@{}}Stiffness of front wheels: $163.9\frac{1}{s^2}$ and of back wheels: $275.4\frac{1}{s^2}$\\Damper front and back is set to $30\frac{1}{s}$. The equations used are inspired by \cite{Suspensions},\\  which may differ to the approach taken in Rocket League that remains unclear.\\ \end{tabular}\\

Car-to-car interaction                &                                                                & Implemented using the Bullet and Psyonix impulses, but not thoroughly tested  
\\
 \rowcolor[HTML]{EFEFEF}
Demolitions                           & \cite{Demolitions}                                                                    & Implemented, but not thoroughly tested and hence not considered in this paper 
\\
\hline
\end{tabular}
\label{tab:physics}
\vspace{-0.15in}
\end{table*}
\renewcommand{\arraystretch}{1}

\begin{figure}
    \centering
    \includegraphics[width=0.45\textwidth]{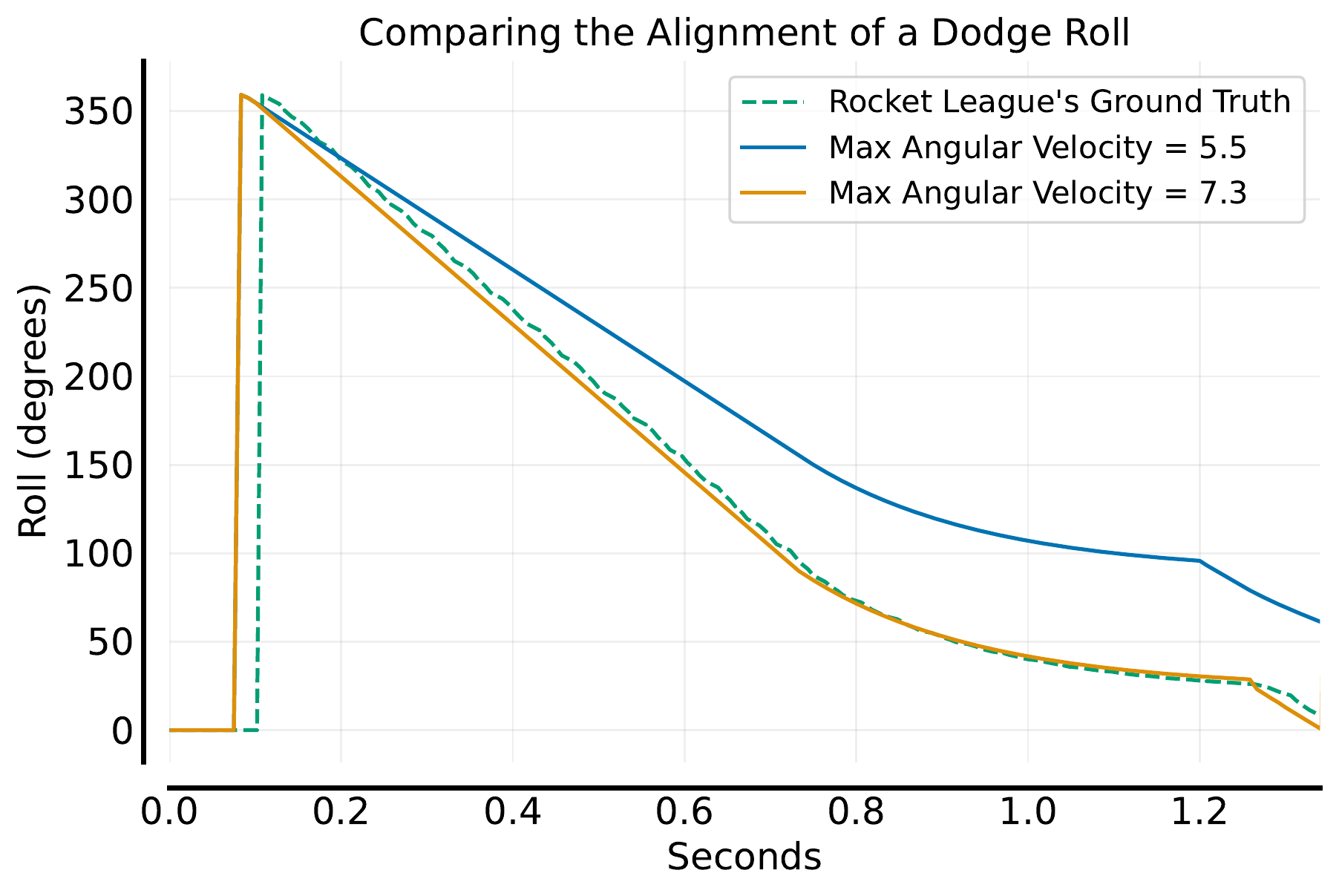}
    \caption{The physical maneuver of a dodge roll is executed to exemplary show the alignment of the Unity simulation to the ground truth by using different max angular velocities.}
    \label{fig:dodge_alignment}
\vspace{-0.15in}
\end{figure}

The implementation of the Unity simulation originates from the so called \emph{RoboLeague} repository \cite{Roboleague2021}.
As this version of the simulation is by far incomplete and inaccurate, multiple fundamental aspects and concepts are implemented, which are essentially based on the physical specifications of Rocket League.
These comprise, for example, the velocity and acceleration of the ball and the car, as well as the concept of boosting.
Jumps, double jumps as well as dodge rolls are now possible, and also collisions and interactions.
There is friction caused by the interaction of a car with the ground, but also friction caused by the air is taken into account.

However, further adjustments are necessary.
Therefore, table \ref{tab:physics} provides an overview of all the material that was considered during implementing essential physical components, while highlighting distinct adjustments that differ from the information provided by the references.
It has to be noted that most measures are given in \emph{unreal units} (uu).
To convert them to Unity's scale, these have to be divided by $100$.

Some adjustments are based on empirical findings by comparing the outcome of distinct physical maneuvers inside the implemented training simulation and the ground truth provided by Rocket League.
A physical maneuver simulates several player inputs over time, such as applying throttle and steering left or right.
While the simulation is conducted in both simulations, multiple relevant game state variables like positions, rotations, and velocities are monitored for later evaluation.
Figure \ref{fig:dodge_alignment} is an example where the physical maneuver orders the car to execute a dodge roll.
Whereas the original max angular velocity of $5.5\rs$ does not compare well to the ground truth, a more suitable value of $7.3 \rs$ is found by analyzing the observed data.

The speed of the training simulation is about $950~steps/second$, while RLBot is constrained to the real-time, where only $120~steps/second $ are possible.
This simulation performance is measured on a Windows Desktop utilizing a GTX 1080 and a AMD Ryzen 7 3700X.

\subsection{Goalie Environment}

In the goalie environment, the agent is asked to save shots.
1000 different samples of shots, which uniformly vary in speed, direction, and origin, are faced by the agent during training.
In every episode, one shot is fired towards the agent's goal.
The agent's position is reset to the center of the goal at the start of each episode.
Every save rewards the agent with $+1$.
A goalkeeping episode terminates if the ball hits the goal or is deflected by the agent.

\subsection{Striker Environment}

To score a goal is the agent's task inside the striker environment.
The ball moves bouncy, slowly, close, and in parallel to the goal.
Its speed and origin are sampled uniformly from 1000 samples during the agent's training.
The agent's position is farther away from the goal while being varied as well.
$+1$ is the only reward signal that the agent receives upon scoring.
Once the ball hits the goal or a time limit is reached, the episode terminates and the environment is reset.

\subsection{Observation and Action Space}

\begin{figure}
	\includegraphics[width=0.49\textwidth]{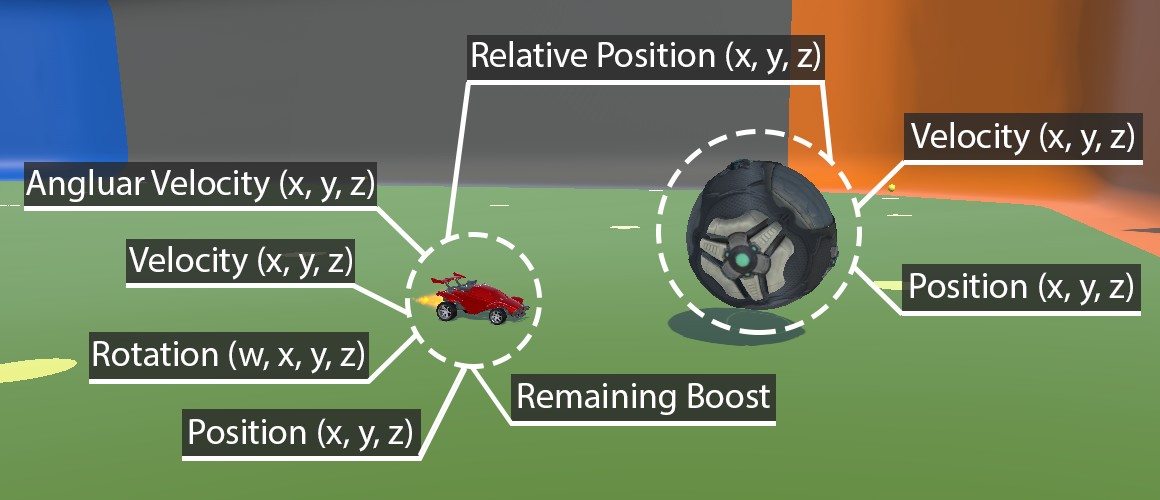}
	\caption{The contents of the agent's observation.}
	\label{img:obs_space}
\vspace{-0.15in}
\end{figure}

Both environments share the same observation and action space.
The agent perceives 23 normalized game state variables to fully observe its environment as illustrated by figure \ref{img:obs_space}.
The agent's action space is multi-discrete and contains the following 8 dimensions:
\vspace{-0.1in}
\begin{itemize}
    \begin{multicols}{2}
    \item Throttle (5 actions)
    \item Steer (5 actions)
    \item Yaw (5 actions)
    \item Pitch (5 actions)
    \item Roll (3 actions)
    \item Boost (2 actions)
    \item Drift or Air Roll (2 actions)
    \item Jump (2 actions)
    \end{multicols}
\end{itemize}
\vspace{-0.1in}
Rocket League is usually played by humans using a gamepad as input device.
Some of the inputs (e.g. thumbstick) are thus continuous and not discrete.
To simplify the action space, the continuous actions throttle, steer, yaw, and pitch are discretized using buckets as suggested by Pleines et al. \cite{Pleines2019}.
By this means, the agent picks one value from a bucket containing the values $-1$, $-0.5$, $0$, $0.5$  and $1$.
The roll action is also discretized using the values $-1$, $0$ and $1$.
All other actions determine whether the concerned discrete action is executed or not.
The action dimension that is in charge of drifting and air rolling is another special case.
Both actions can be boiled down to one because drifting is limited to being on the ground, whereas air rolling can be done in the air only.
Moreover, multi-discrete action spaces allow the execution of concurrent actions.
One discrete action dimension could achieve the same behavior.
This would require defining actions that feature every permutation of the available actions.
As a consequence, the already high-dimensional action space of Rocket League would be much larger and therefore harder to train.

\section{Deep Reinforcement Learning}

The actor-critic, on-policy algorithm PPO~\cite{Schulman2017} and its clipped surrogate objective (Equation \ref{eq:ppo}) is used to train the agent's policy $\pi$, with respect to its model parameters $\theta$, inside the Unity simulation.
PPO, algorithmic details, and the model architecture are presented next.

\subsection{Proximal Policy Optimization}

$L_t^C(\theta)$ denotes the policy objective, which optimizes the probability ratio of the current policy $\pi_\theta$ and the old one $\pi_{\theta old}$:
\small
\begin{equation}
    L^{C}_t(\theta) = \hat{\mathbb{E}}_t [min(q_t(\theta)\hat{A}_t,clip(q_t(\theta),1- \epsilon,1+\epsilon)\hat{A}_t)]
    \label{eq:ppo}
\end{equation}
\begin{equation*}
    \textnormal{with the surrogate objective}~q_t(\theta) = \frac{\pi _\theta(a_t|s_t)}{\pi _{\theta \text{old}}(a_t|s_t)}
\end{equation*}
\normalsize
$s_t$ is the environment's state at step $t$.
$a_t$ is an action tuple, which is executed by the agent, while being in $s_t$.
The clipping range is stated by $\epsilon$ and $\hat{A}_t$ is the advantage, which is computed using generalized advantage estimation~\cite{Schulman2015GAE}.
While computing the squared error loss $L_t^V$ of the value function, the maximum between the default and the clipped error loss is determined. 
\small
\begin{equation}
        V_t^{C} = V_{\theta old}(s_t) + clip(V_\theta(s_t) - V_{\theta old}(s_t), -\epsilon, \epsilon)
\end{equation}
\begin{equation}
    L_t^V = max((V_\theta(s_t) - G_t)^2, (V_t^C-G_t)^2)
\end{equation}
\begin{equation*}
    \textnormal{with the sampled return}~G_t = V_{\theta old}(s_t) + \hat{A}_t
\end{equation*}
\normalsize
The final objective is established by $L^{CVH}_t(\theta)$:
\small
\begin{equation}
	L^{CVH}_t(\theta)=\hat{\mathbb{E}}_t [L^{C}_t(\theta)-c_1L^{V}_t(\theta)+c_2\mathcal{H}[\pi_\theta](s_t)]
\end{equation}
\normalsize
To encourage exploration, the entropy bonus $\mathcal{H}[\pi_\theta](s_t)$ is added and weighted by the coefficient $c_2$.
Weighting is also applied to the value loss using $c_1$.

\subsection{Algorithmic Details and Model Architecture}

PPO starts out by sampling multiple trajectories of experiences, which may contain multiple completed and truncated episodes, from a constant number of concurrent environments (i.e. workers).
The model parameters are then optimized by conducting stochastic gradient descent for several epochs of mini-batches, which are sampled from the collected data.
Before computing the loss function, advantages are normalized across each mini-batch.
The computed gradients are clipped based on their norm.

\begin{figure}
    \centering
	\includegraphics[width=0.375\textwidth]{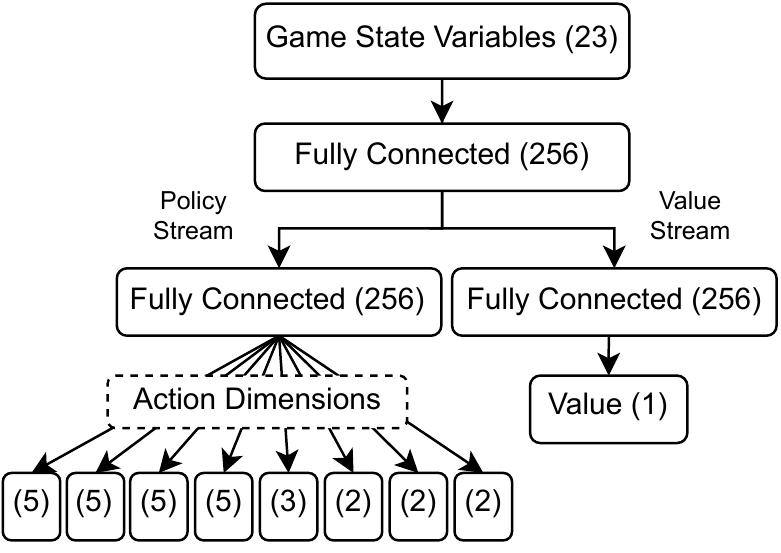}
	\caption{The policy and the value function share gradients and several parameters.
	After feeding 23 game states variables as input to the model and processing a shared fully connected layer, the network is split into a policy and value stream starting with their own fully connected layer.
	The policy stream outputs action probabilities for each available action dimension, whereas the value stream exposes its estimated state-value.}
	\label{img:model}
\vspace{-0.15in}
\end{figure}

A relatively shallow neural net (model) is shared by the value function and the policy (Figure \ref{img:model}).
To support multi-discrete actions, the policy head of the model outputs 8 categorical action probability distributions.
During action selection, each distribution is used to sample actions, which are provided to the agent as a tuple.
The only adjustment to the policy's loss computation is that the probabilities of the selected actions are concatenated.
Concerning the entropy bonus, the mean of the action distributions' entropies is used.


\begin{table*}
\caption{The resulted error for each run physical maneuver scenario. The car's position is considered by the green shaded data points, while the blue ones are related to the ball's position.}
\centering
\resizebox{\textwidth}{!}{
\begin{tabular}{l|
>{\columncolor[HTML]{F1FFF0}}r 
>{\columncolor[HTML]{F1FFF0}}r 
>{\columncolor[HTML]{F1FFF0}}r |
>{\columncolor[HTML]{F1FFF0}}r 
>{\columncolor[HTML]{F1FFF0}}r 
>{\columncolor[HTML]{F1FFF0}}r |
>{\columncolor[HTML]{F1FFF0}}r 
>{\columncolor[HTML]{F1FFF0}}r 
>{\columncolor[HTML]{F1FFF0}}r |
>{\columncolor[HTML]{F1FFF0}}r 
>{\columncolor[HTML]{F1FFF0}}r 
>{\columncolor[HTML]{F1FFF0}}r |
>{\columncolor[HTML]{EEEFFF}}r 
>{\columncolor[HTML]{EEEFFF}}r 
>{\columncolor[HTML]{EEEFFF}}r |
>{\columncolor[HTML]{EEEFFF}}r 
>{\columncolor[HTML]{EEEFFF}}r 
>{\columncolor[HTML]{EEEFFF}}r |}
\cline{2-19}

& \multicolumn{3}{c|}{\cellcolor[HTML]{F1FFF0}\textbf{1) Acceleration}}                                                    & \multicolumn{3}{c|}{\cellcolor[HTML]{F1FFF0}\textbf{2) Air Control}}                                                     & \multicolumn{3}{c|}{\cellcolor[HTML]{F1FFF0}\textbf{3) Drift}}                                                             & \multicolumn{3}{c|}{\cellcolor[HTML]{F1FFF0}\textbf{4) Jump}}                                                            & \multicolumn{3}{c|}{\cellcolor[HTML]{EEEFFF}\textbf{5) Ball Bounce}}                                                     & \multicolumn{3}{c|}{\cellcolor[HTML]{EEEFFF}\textbf{6) Shot}}                                                               \\ \hline
\multicolumn{1}{|l|}{\cellcolor[HTML]{F4F4F4}\textbf{Mean} } & \multicolumn{1}{r|}{\cellcolor[HTML]{F1FFF0}0.69} & \multicolumn{1}{r|}{\cellcolor[HTML]{F1FFF0}3.72} & 1.67 & \multicolumn{1}{r|}{\cellcolor[HTML]{F1FFF0}2.32} & \multicolumn{1}{r|}{\cellcolor[HTML]{F1FFF0}3.07} & 5.24  & \multicolumn{1}{r|}{\cellcolor[HTML]{F1FFF0}3.19}  & \multicolumn{1}{r|}{\cellcolor[HTML]{F1FFF0}0.84} & 4.87  & \multicolumn{1}{r|}{\cellcolor[HTML]{F1FFF0}0.07} & \multicolumn{1}{r|}{\cellcolor[HTML]{F1FFF0}0.22} & 1.37 & \multicolumn{1}{r|}{\cellcolor[HTML]{EEEFFF}0.01} & \multicolumn{1}{r|}{\cellcolor[HTML]{EEEFFF}0.05} & 0.03 & \multicolumn{1}{r|}{\cellcolor[HTML]{EEEFFF}28.45} & \multicolumn{1}{r|}{\cellcolor[HTML]{EEEFFF}23.79} & 28.31 \\ \hline
\multicolumn{1}{|l|}{\cellcolor[HTML]{F4F4F4}\textbf{Std}}        & \multicolumn{1}{r|}{\cellcolor[HTML]{F1FFF0}0.48} & \multicolumn{1}{r|}{\cellcolor[HTML]{F1FFF0}3.05} & 1.92 & \multicolumn{1}{r|}{\cellcolor[HTML]{F1FFF0}2.61} & \multicolumn{1}{r|}{\cellcolor[HTML]{F1FFF0}1.81} & 4.58  & \multicolumn{1}{r|}{\cellcolor[HTML]{F1FFF0}4.73}  & \multicolumn{1}{r|}{\cellcolor[HTML]{F1FFF0}1.04} & 6.80   & \multicolumn{1}{r|}{\cellcolor[HTML]{F1FFF0}0.06} & \multicolumn{1}{r|}{\cellcolor[HTML]{F1FFF0}0.15} & 0.69 & \multicolumn{1}{r|}{\cellcolor[HTML]{EEEFFF}0.01} & \multicolumn{1}{r|}{\cellcolor[HTML]{EEEFFF}0.03} & 0.02 & \multicolumn{1}{r|}{\cellcolor[HTML]{EEEFFF}25.20}  & \multicolumn{1}{r|}{\cellcolor[HTML]{EEEFFF}25.01} & 22.49 \\ \hline
\multicolumn{1}{|l|}{\cellcolor[HTML]{F4F4F4}\textbf{Max}}  & \multicolumn{1}{r|}{\cellcolor[HTML]{F1FFF0}1.21} & \multicolumn{1}{r|}{\cellcolor[HTML]{F1FFF0}8.04} & 5.96 & \multicolumn{1}{r|}{\cellcolor[HTML]{F1FFF0}8.40}  & \multicolumn{1}{r|}{\cellcolor[HTML]{F1FFF0}8.12} & 12.97 & \multicolumn{1}{r|}{\cellcolor[HTML]{F1FFF0}16.06} & \multicolumn{1}{r|}{\cellcolor[HTML]{F1FFF0}5.12} & 21.08 & \multicolumn{1}{r|}{\cellcolor[HTML]{F1FFF0}0.24} & \multicolumn{1}{r|}{\cellcolor[HTML]{F1FFF0}0.41} & 2.02 & \multicolumn{1}{r|}{\cellcolor[HTML]{EEEFFF}0.02} & \multicolumn{1}{r|}{\cellcolor[HTML]{EEEFFF}0.12} & 0.07 & \multicolumn{1}{r|}{\cellcolor[HTML]{EEEFFF}58.16} & \multicolumn{1}{r|}{\cellcolor[HTML]{EEEFFF}59.00}    & 58.19 \\ \hline
\end{tabular}
}
\label{tab:align}
\vspace{-0.15in}
\end{table*}

\section{Sim-to-sim Transfer}
\label{sec:sim-to-sim}

Two major approaches are considered to examine learned behaviors inside the Unity simulation and its transfer to Rocket League.
The first one runs various handcrafted scenarios (like seen in section \ref{sec:env}) in both simulations to directly compare their alignment.
This way, it can be determined whether the car or the ball behave similarly or identically concerning their positions and velocities.
The second approach trains the agent in Unity given the goalie and the striker environment, while all implemented physics components are included.
We further conduct an ablation study on the implemented physics where each experiment turns off one or all components.
Turning off may also refer to use the default physics of Unity.

If not stated otherwise, each training run is repeated 5 times and undergoes a thorough evaluation.
Each model checkpoint is evaluated in Unity and Rocket League by 10 training and 10 novel shots, which are repeated 3 times.
Therefore, each data point aggregates 150 episodes featuring one shot.
Result plots show the interquartile mean of the cumulative reward and a confidence interval of 95\% as recommended by Agarwal et al. \cite{agarwal2021statistics}.
The hyperparameters are detailed in Table \ref{tab:hyperparameters}.
At last, we describe some of the learned behaviors that are also retrieved from training in a more difficult striker environment.

\begin{table}[]
\centering
\caption{The hyperparameters used to conduct the training with PPO. The learning rate $\alpha$ and $c_2$ decay linearly over time.}
\begin{tabular}{l|r|l|r}
\rowcolor[HTML]{EFEFEF} 
\multicolumn{1}{l|}{\cellcolor[HTML]{EFEFEF}Hyperparameter} & \multicolumn{1}{c|}{\cellcolor[HTML]{EFEFEF}Value} & \multicolumn{1}{l|}{\cellcolor[HTML]{EFEFEF}Hyperparameter} & \multicolumn{1}{c}{\cellcolor[HTML]{EFEFEF}Value} \\ \hline
Discount Factor $\gamma$                                             & 0.99                                               & Clip Range $\epsilon$                                            & 0.2                                               \\
\rowcolor[HTML]{EFEFEF} 
$\lambda$ (GAE)                                                  & 0.95                                               & $c_1$                                       & 0.25                                              \\
Number of Workers                                            & 16                                                 & Initial $\alpha$                                                & 0.0003                                            \\
\rowcolor[HTML]{EFEFEF} 
Worker Steps                                                 & 512                                                & Min $\alpha$                                                    & 0.000003                                          \\
\rowcolor[HTML]{FFFFFF} 
Batch Size                                                   & 8192                                               & Initial $c_2$                                                & 0.0005                                            \\
\rowcolor[HTML]{EFEFEF} 
Epochs                                                       & 3                                                  & Min $c_2$                                                      & 0.00001                                           \\
\rowcolor[HTML]{FFFFFF} 
Mini Batches                                                 & 8                                                  & Optimizer                                                    & AdamW                                             \\
\rowcolor[HTML]{EFEFEF} 
Max Gradient Norm                                                   & 0.5                                                & Activations                                                  & ReLU                                             
\end{tabular}
\label{tab:hyperparameters}
\vspace{-0.15in}
\end{table}

\subsection{Alignment Comparison using Handcrafted Scenarios}
\label{sec:alginment}


To directly compare the alignment between both simulations, six physical maneuvers are assessed by 3 different handcrafted scenarios:
\begin{enumerate}
    \item Acceleration
    \begin{itemize}
        \item Car drives forward and steers left and right
        \item Car drives backward and steers left and right
        \item Car uses boost and steers left and right
    \end{itemize}
    \item Air Control
    \begin{itemize}
        \item Car starts up in the air, looks straight up, boosts shortly and boosts while rolling in the air
        \item Car starts up in the air, has an angle of $45^{\circ}$, boosts shortly and boosts while rolling in the air
        \item Car starts up in the air, looks straight up and concurrently boosts, yaws, and air rolls
    \end{itemize}
    \item Drift
    \begin{itemize}
        \item Car drives forward for a bit and then starts turning and drifting while moving forward
        \item Car drives backward for a bit and then starts turning and drifting while moving forward
        \item Car uses boost and then starts turning and drifting while using boost
    \end{itemize}
    \item Jump
    \begin{itemize}
        \item Car makes a short jump, then a long one and at last a double jump
        \item Car makes a front flip, a back flip and a dodge roll
        \item Car drives forward, does a diagonal front flip and at last a back flip
    \end{itemize}
    \item Ball Bounce
    \begin{itemize}
        \item Ball falls straight down
        \item Ball falls down with an initial force applied on its x-axis
        \item Ball falls down with an initial force applied on its x-axis and an angular velocity
    \end{itemize}
    \item Shot
    \begin{itemize}
        \item Car drives forward and hits the motionless ball
        \item Car drives forward and the ball rolls to the car
        \item Ball jumps, the car jumps while boosting and hits the ball using a front flip
    \end{itemize}
\end{enumerate}
Each scenario tracks the position of the ball and the car during each frame.
As both simulations end up monitoring the incoming data with slight time differences, the final data is interpolated to match in shape.
Afterward, the error for each data point between both simulations is measured.
The final results are described by Table \ref{tab:align}, which comprises the mean, max, and standard deviation (Std) error across each run scenario.
Letting the ball bounce for some time shows the least error, while a significant one is observed when examining the scenarios where the car shoots the ball.
Note that slight inaccuracies during acceleration may cause a strongly summed error when considering a different hit location on the ball.
The other scenarios, where the error is based on the car's position, also indicate that the Unity simulation suffers from inaccuracies.

\subsection{Physics Ablation Study based on PPO Training}

\begin{figure*}
\centering
\subfigure{\includegraphics[clip, trim=0 0 0 0, width=0.255\textwidth]{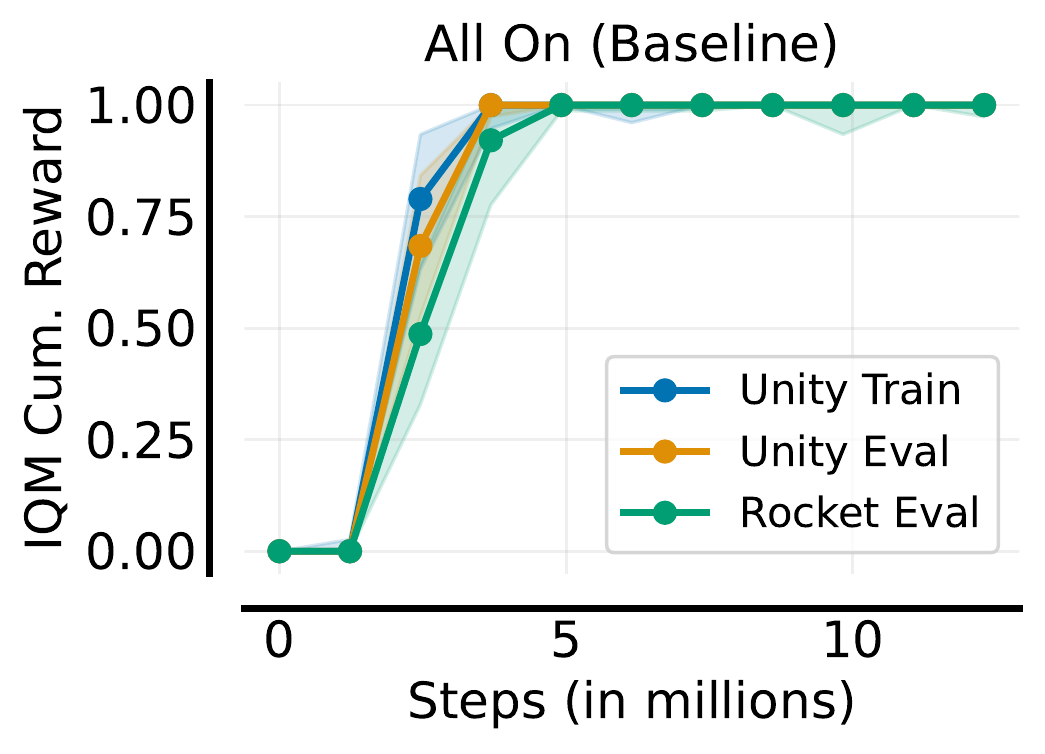}}
\subfigure{\includegraphics[clip, trim=0.7cm 0 0 0, width=0.2375\textwidth]{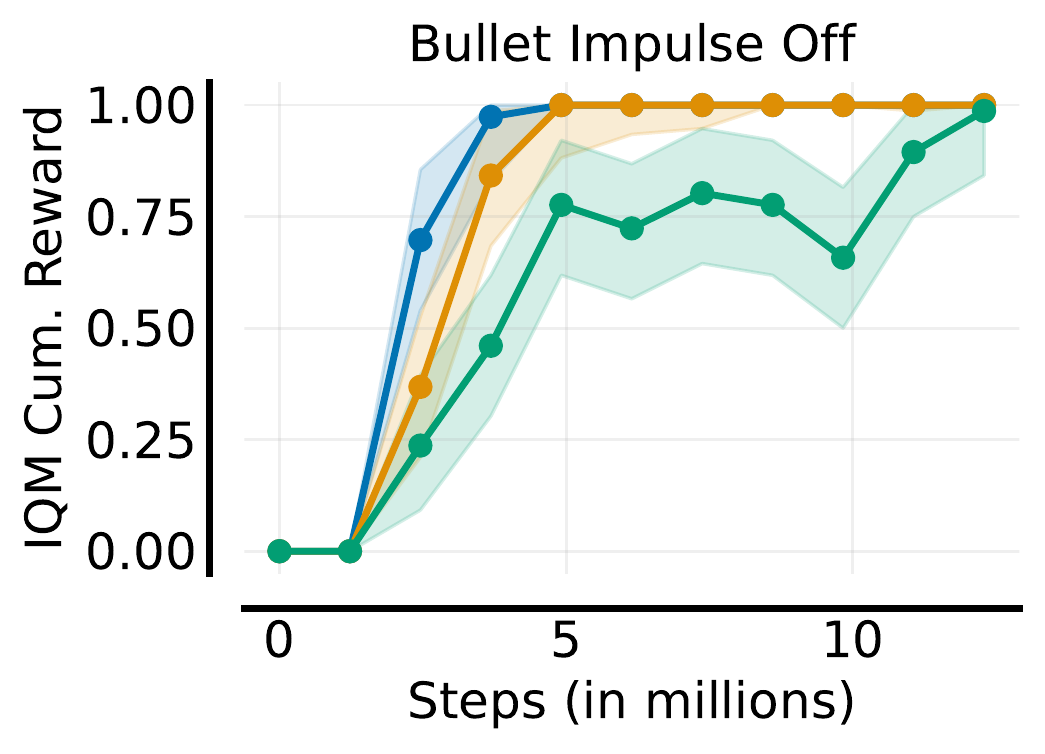}}
\subfigure{\includegraphics[clip, trim=0.7cm 0 0 0, width=0.2375\textwidth]{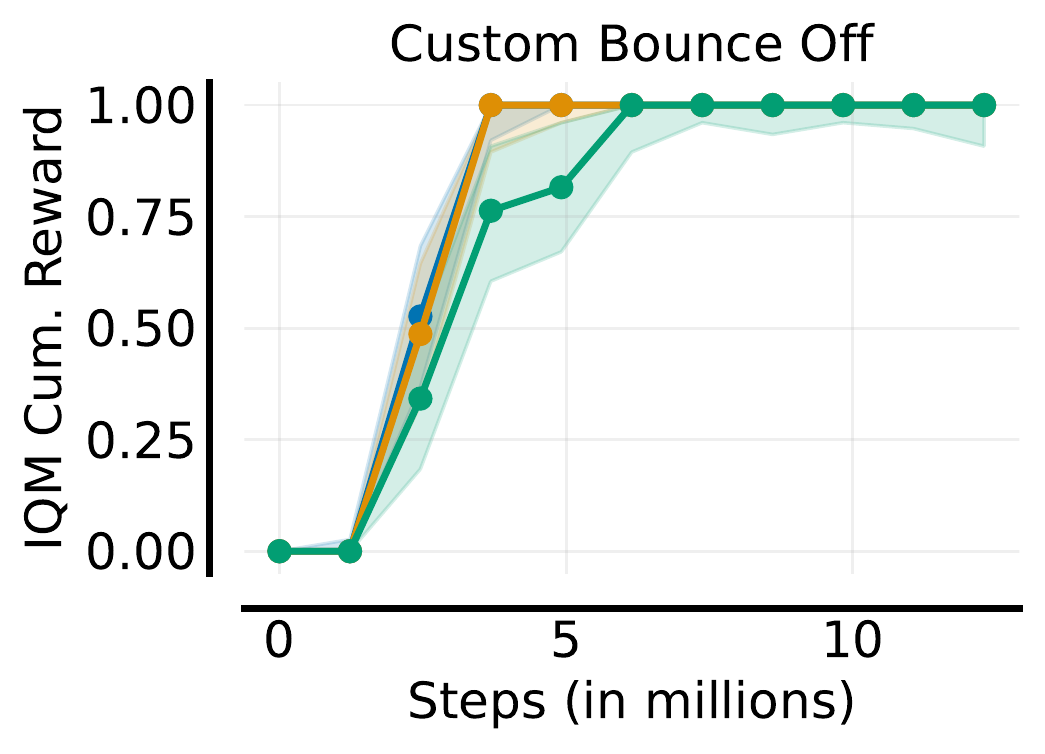}}
\subfigure{\includegraphics[clip, trim=0.7cm 0 0 0, width=0.2375\textwidth]{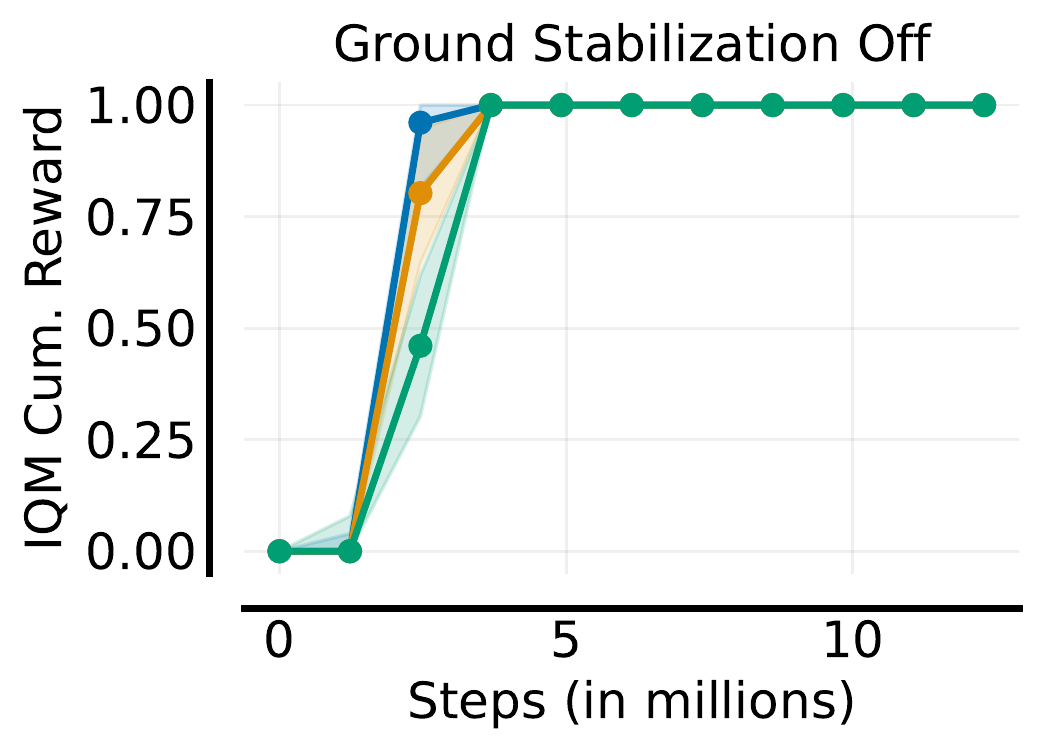}}\\
\vspace{-0.75cm}
\subfigure{\includegraphics[clip, trim=0 0 0 0,width=0.255\textwidth]{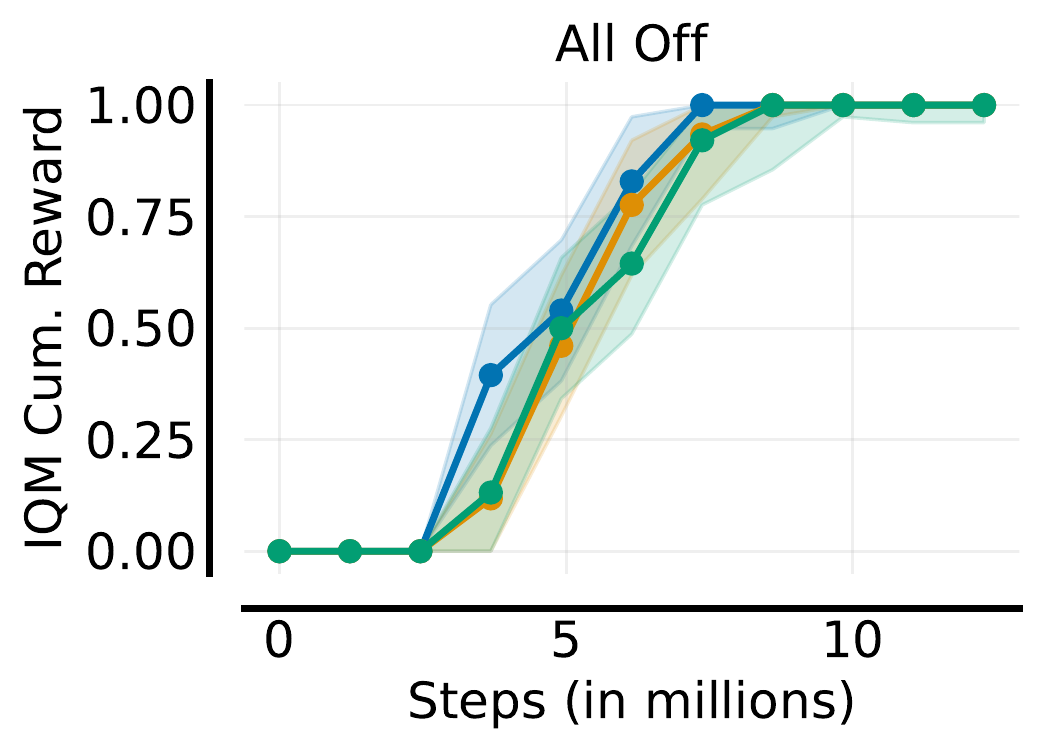}}
\subfigure{\includegraphics[clip, trim=0.7cm 0 0 0,width=0.2375\textwidth]{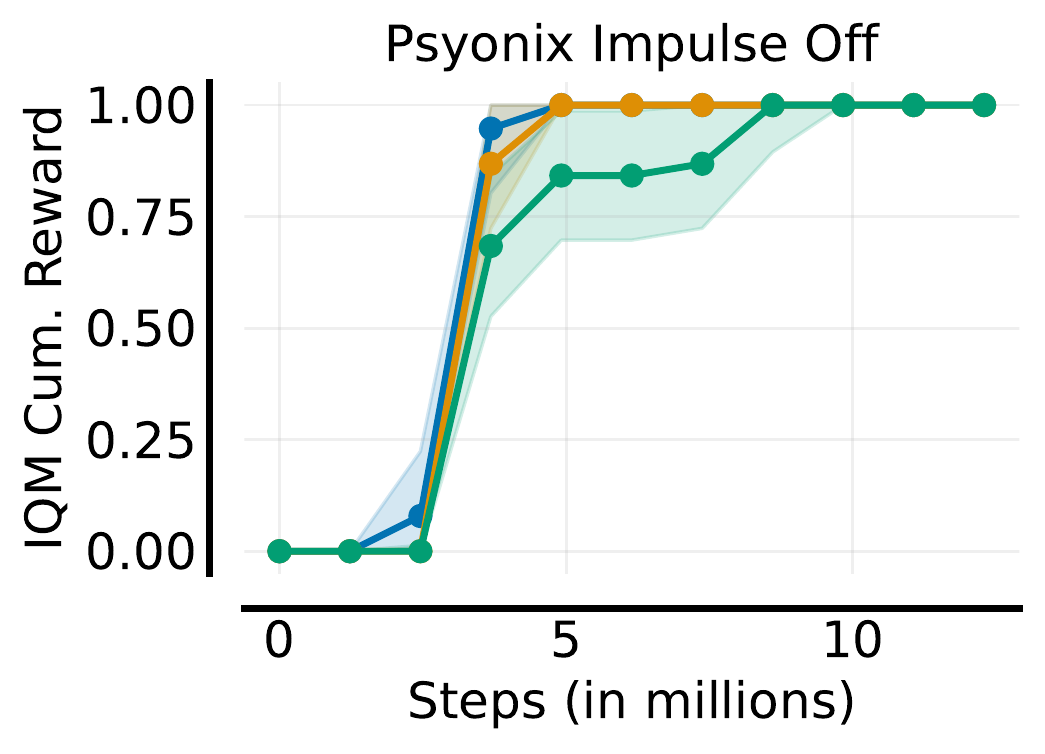}}
\subfigure{\includegraphics[clip, trim=0.7cm 0 0 0,width=0.2375\textwidth]{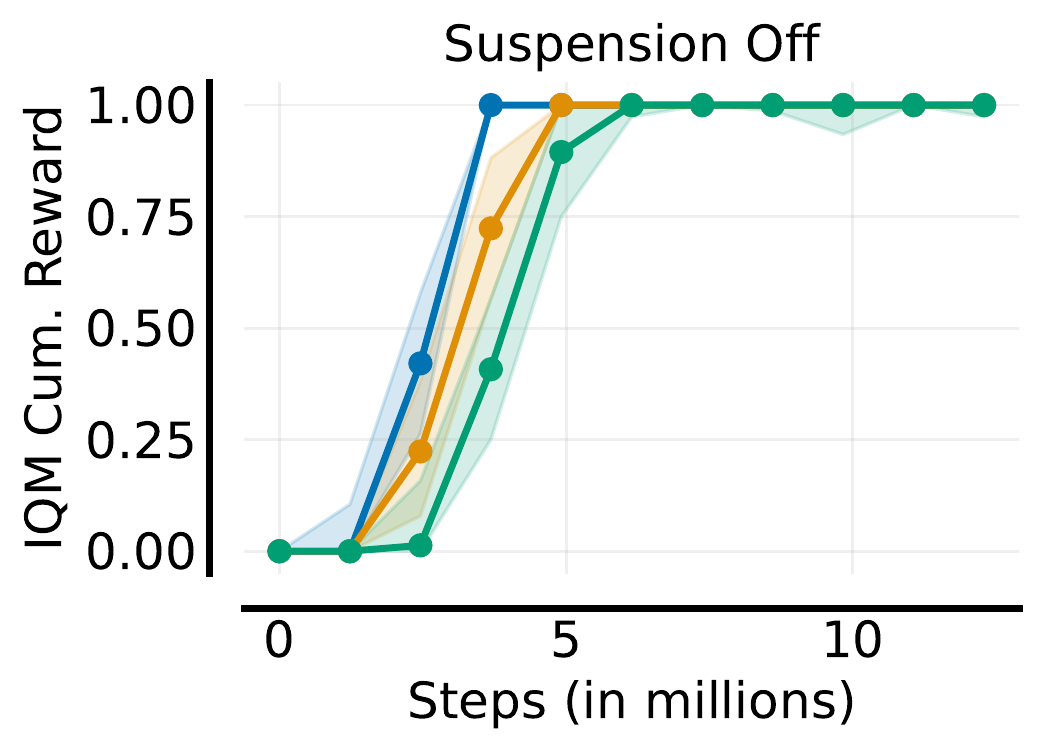}}
\subfigure{\includegraphics[clip, trim=0.7cm 0 0 0,width=0.2375\textwidth]{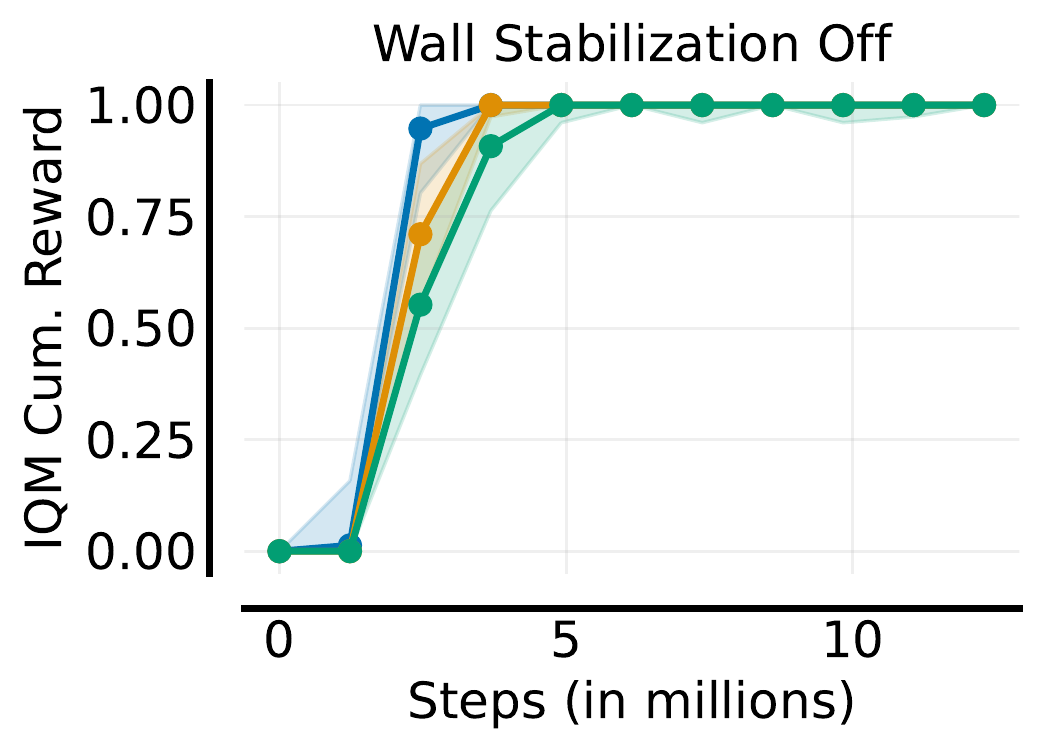}}
\caption{Results of training the goalie environment under different ablations and transferring it to Rocket League. The agent is evaluated on training shots and ones, which were not seen during training. The agent easily solves the goalie task under all circumstances. Both, training and unseen shots, behave identically in Rocket League.}
\label{fig:goalie_results}
\vspace{-0.15in}
\end{figure*}

\begin{figure*}
\centering
\subfigure{\includegraphics[clip, trim=0 0 0 0, width=0.255\textwidth]{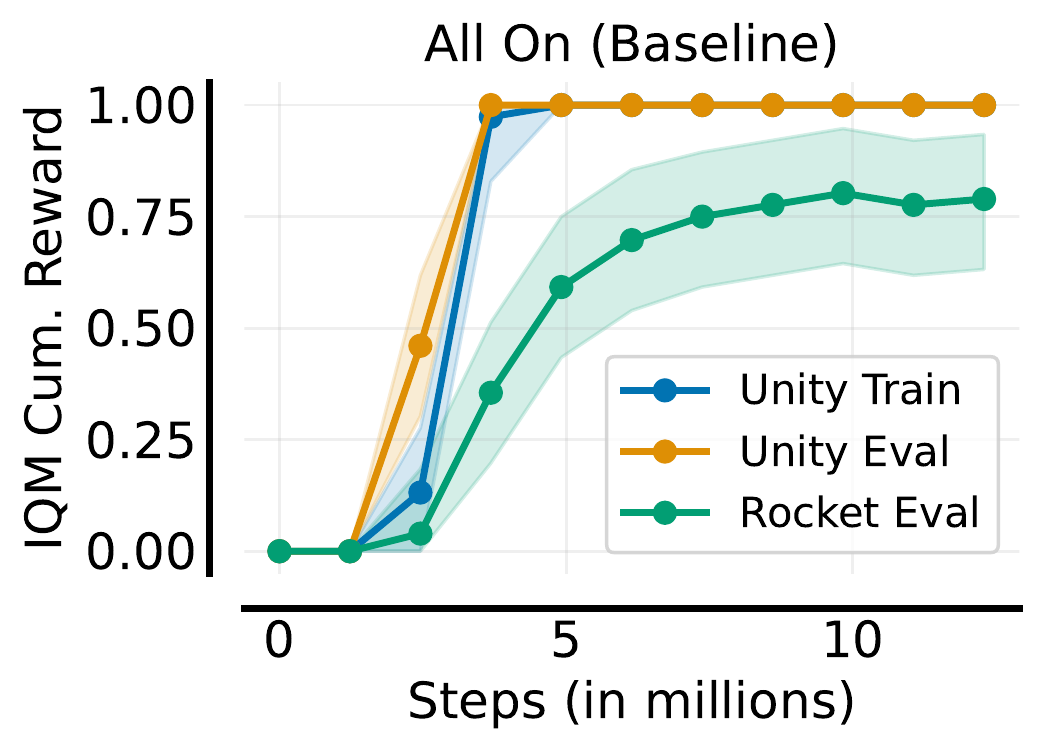}}
\subfigure{\includegraphics[clip, trim=0.7cm 0 0 0, width=0.2375\textwidth]{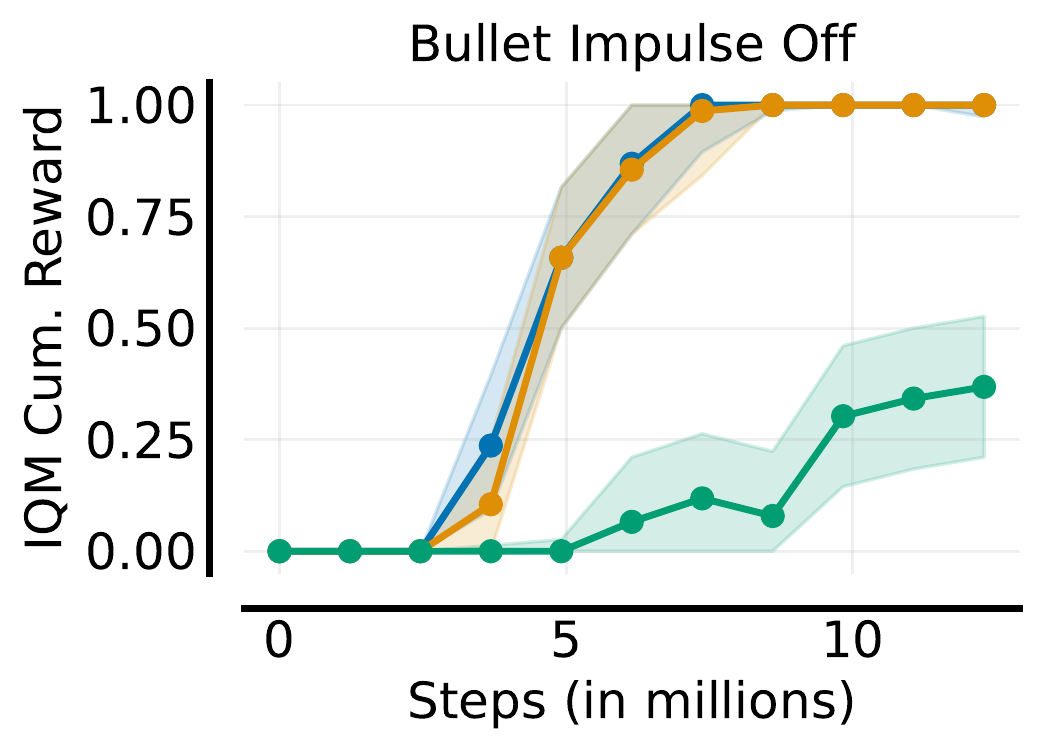}}
\subfigure{\includegraphics[clip, trim=0.7cm 0 0 0, width=0.2375\textwidth]{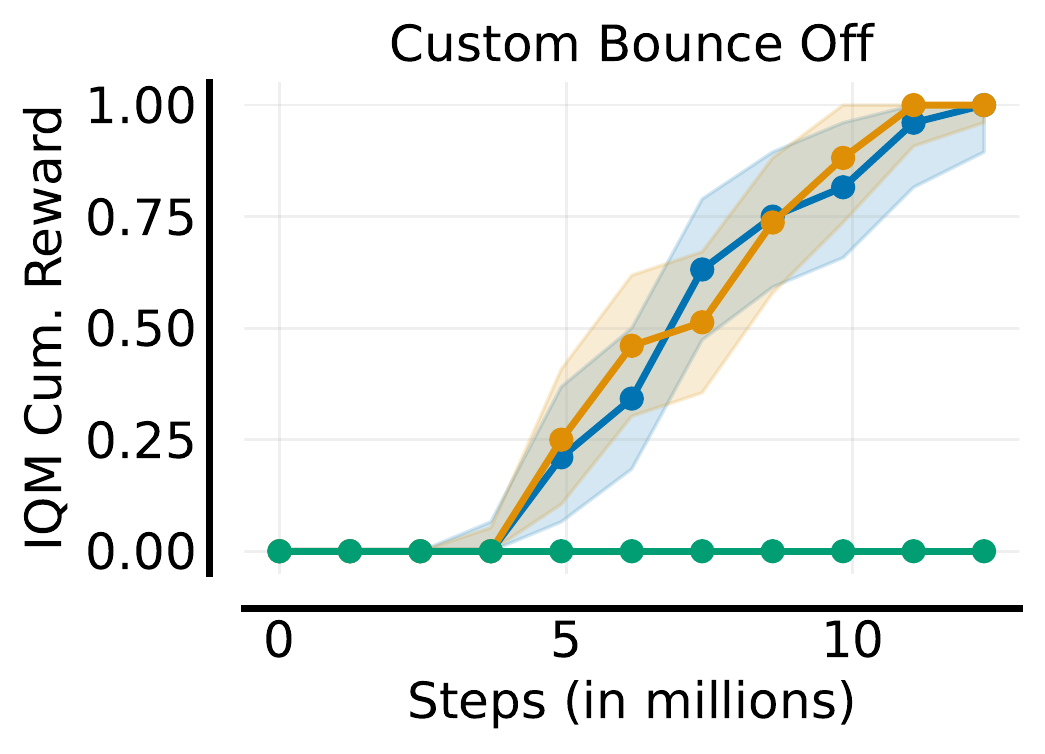}}
\subfigure{\includegraphics[clip, trim=0.7cm 0 0 0, width=0.2375\textwidth]{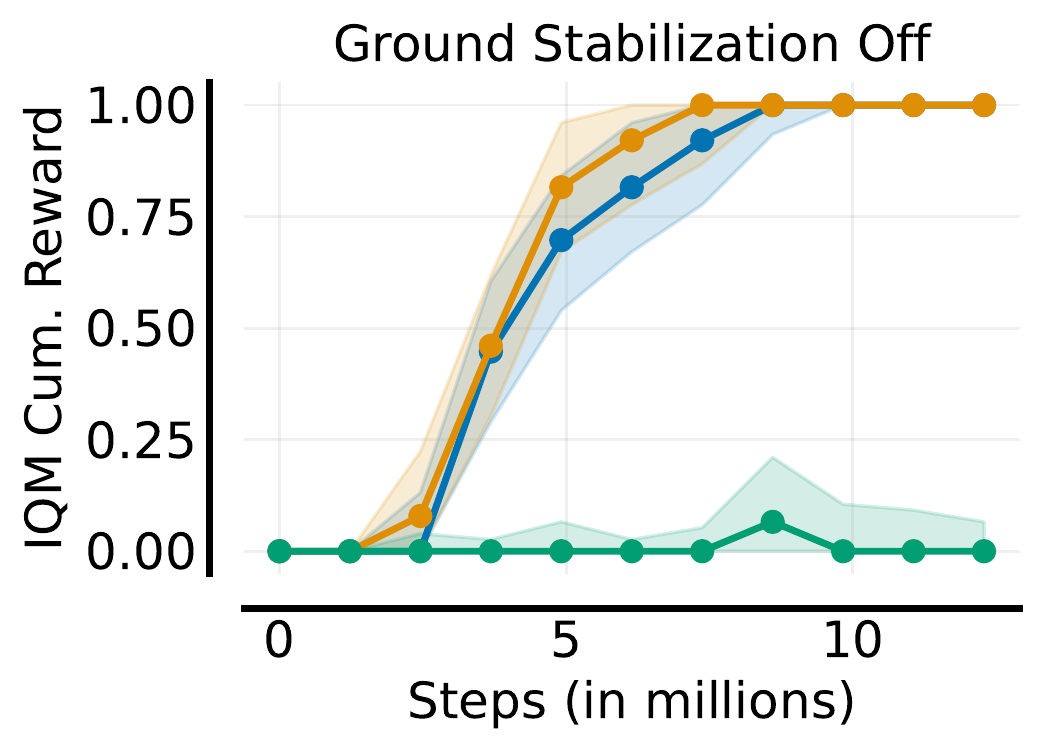}}\\
\vspace{-0.75cm}
\subfigure{\includegraphics[clip, trim=0 0 0 0,width=0.255\textwidth]{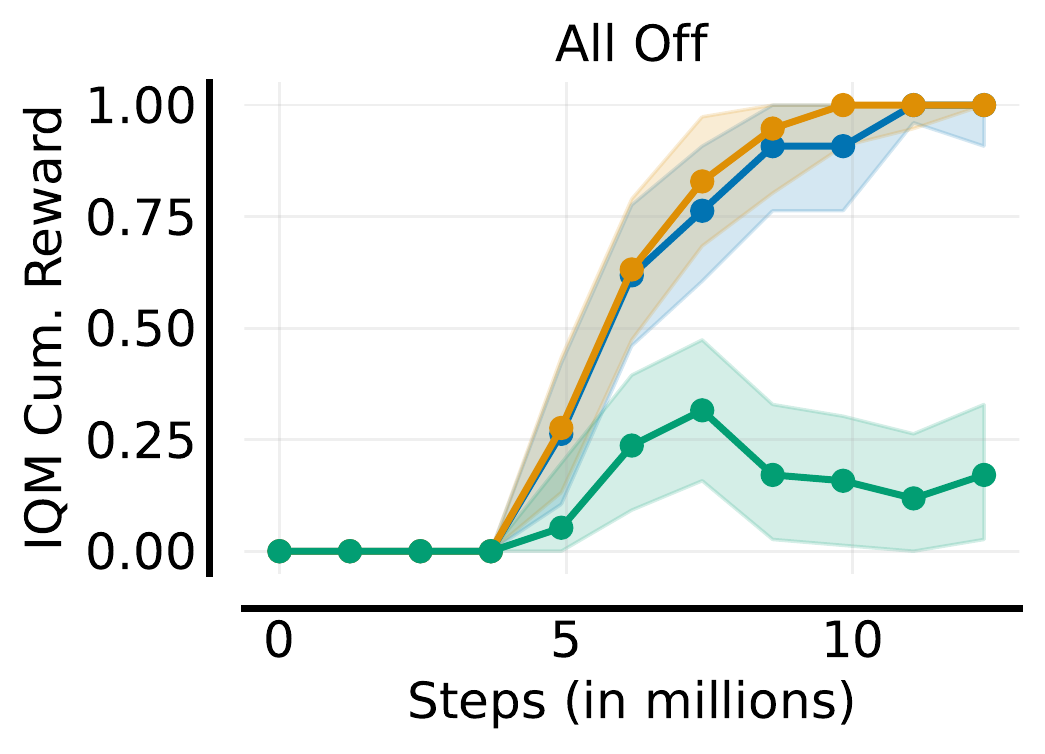}}
\subfigure{\includegraphics[clip, trim=0.7cm 0 0 0,width=0.2375\textwidth]{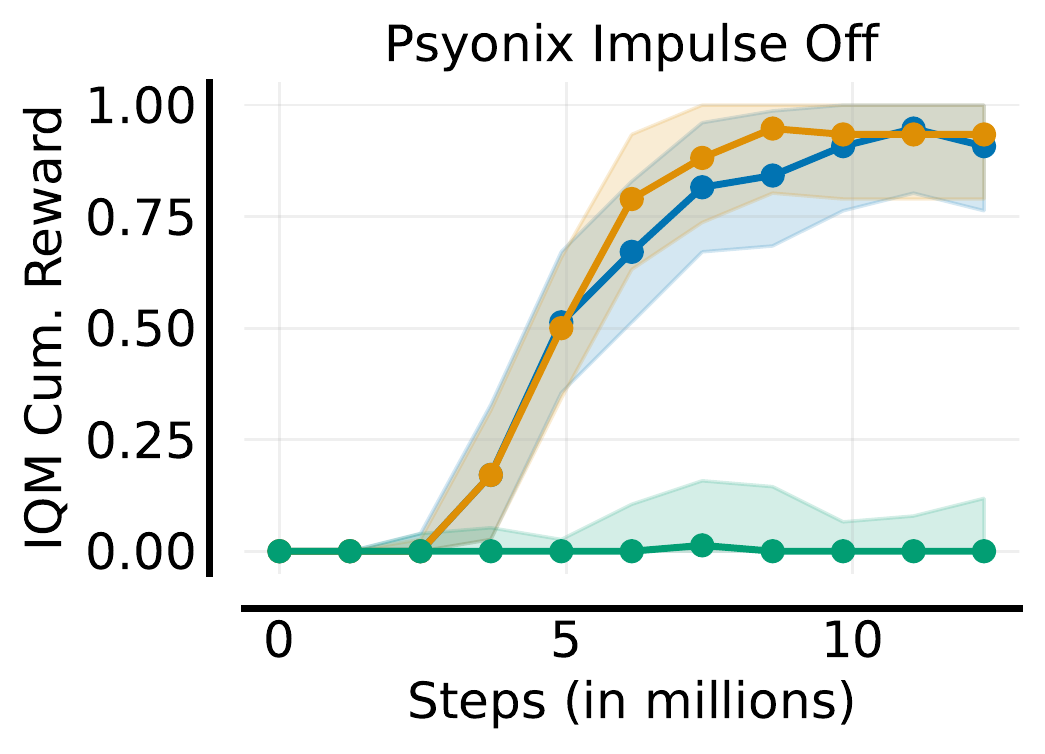}}
\subfigure{\includegraphics[clip, trim=0.7cm 0 0 0,width=0.2375\textwidth]{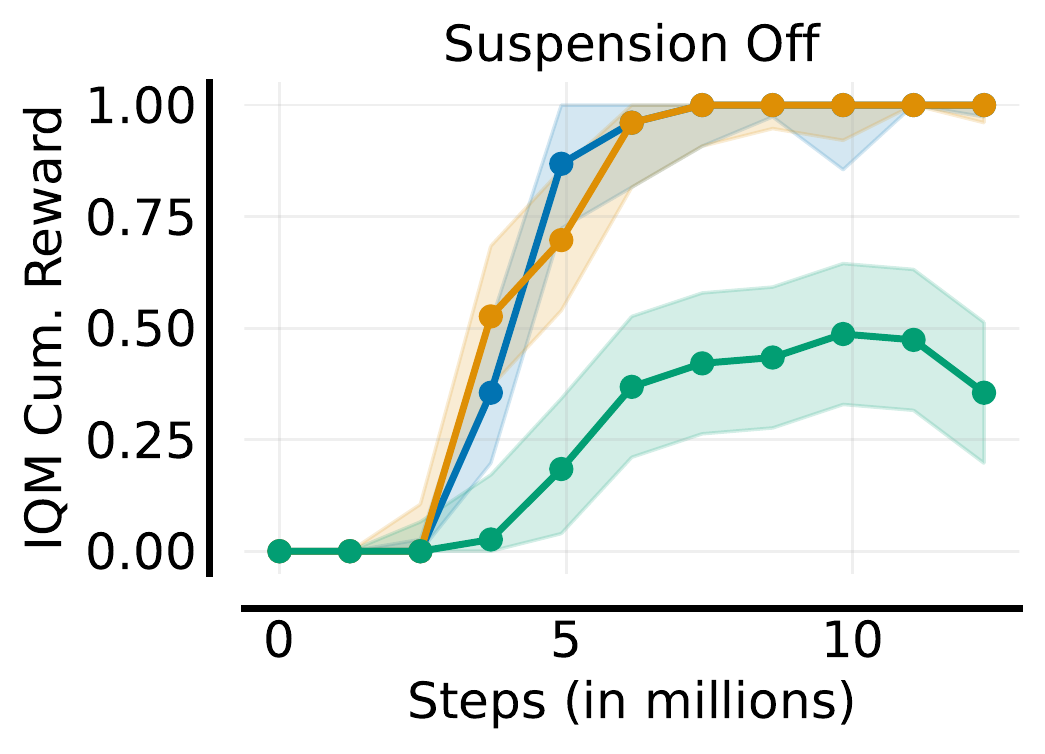}}
\subfigure{\includegraphics[clip, trim=0.7cm 0 0 0,width=0.2375\textwidth]{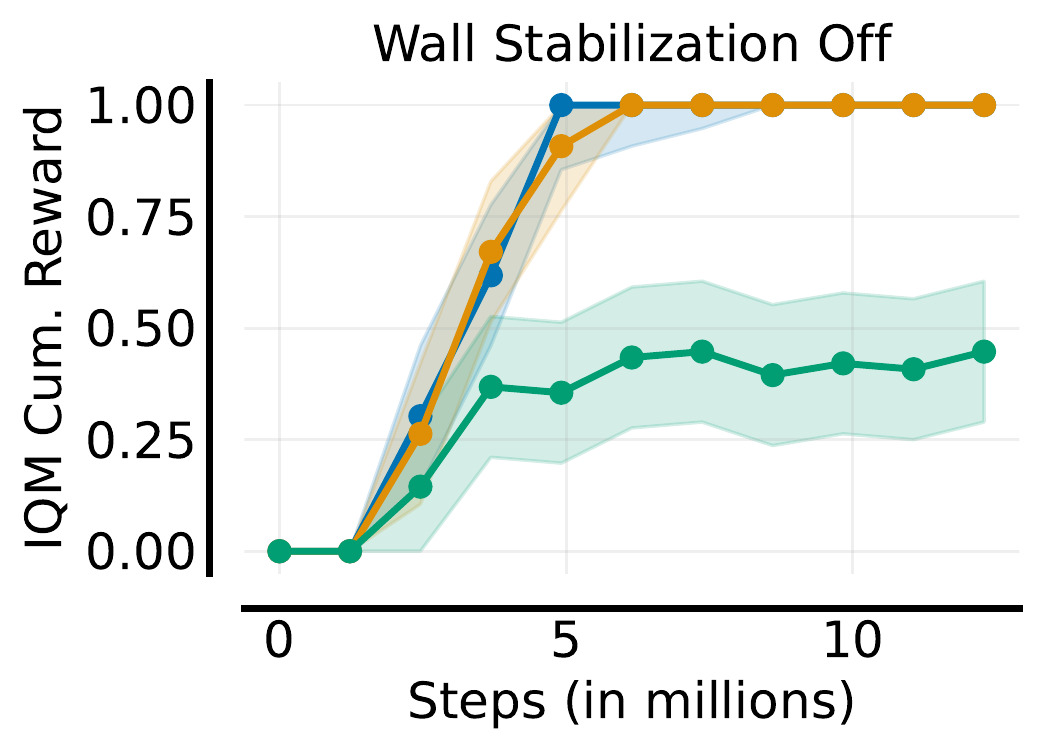}}
\caption{Results of training the striker environment under different ablations and transferring it to Rocket League. The agent is evaluated on training situations and ones, which were not seen during training. The agent scores in about 75\% of the played episodes given all physical adaptations, while any ablation turns out catastrophic. Both, training and unseen situations, behave identically in Rocket League.}
\label{fig:striker_results}
\vspace{-0.15in}
\end{figure*}
        
The previously shown imperfections of the Unity simulation may lead to the impression that successfully transferring a trained behavior is rather unlikely.
This assumption can be negated by considering the results retrieved from training the agent in the goalie environment (Figure \ref{fig:goalie_results}).
Even though each experiment ablates all, single or no physical adaptations, the agent is still capable of saving nearly every ball once transferred to Rocket League.
A drawback of the goalie environment lies in its simplicity because the agent only has to somehow hit the ball to effectively deflect it.
The next step of complexity is posed by the striker environment, where the agent has to land a more accurate hit on the ball to score a goal.
Figure \ref{fig:striker_results} illustrates the results of the striker training.
Notably, when all physical adaptations are present, the transferred behavior manages to score in about 75\% of the played episodes.
Catastrophic performances emerge in Rocket League once single physical adaptations are turned off.

\subsection{Learned Policies}

During the performed experiments, several intriguing agent behaviors emerged\footnote{\url{https://www.youtube.com/watch?v=WXMHJszkz6M&list=PL2KGNY2Ei3ix7Vr_vA-ZgCyVfOCfhbX0C}}.
When trained as a goalkeeper, the agent tries to hit the ball very early, while making its body as big as possible towards the ball.
This is achieved by simultaneously jumping and rolling forward or executing a forward flip.
Concerning the striker environment, the agent usually approaches the ball using its boost.
To get a better angle to the ball, the agent steers left and right or vice versa.
Drifting is sometimes used to aid this purpose.
Jumping is always used when needed.
This is usually the case if the agent is close to the ball, which is located above the agent.
Otherwise, the agent's preference is to stay on the ground.
Further training experiments were conducted in a more difficult striker environment.
The ball is not anymore simply passed in parallel and close to the goal.
Instead, the ball bounces higher and farther away from the goal, which increases the challenge of making a good touch on the ball to score.
Given this setting, two different policies were achieved.
One policy approaches the ball as fast as possible while using a diagonal dodge roll to make the final touch to score.
However, this behavior fails a few shots.
The other emerged behavior can be considered as the opposite.
Depending on the distance and the height of the ball, the agent waits some time or even backs up to ensure that it will hit the ball while being on the ground.
Therefore, the agent avoids jumping.
This is surprising because the agent should maximize its discounted cumulative reward and therefore finish the episode faster.
Although the increased difficulty led to different behaviors, the agent may struggle a lot to get there.
Usually, 2 out of 5 training runs succeeded, while the other ones utterly failed.

\section{Discussion}

In this work, the agent is trained on isolated tasks, which are quite apart from a complete match of Rocket League.
To train multiple cooperative and competitive agents, the first obstacle that comes to mind is the tremendously high computational complexity, which might be infeasible for smaller research groups.
But before going this far, several aspects need to be considered that can be treated in isolation as well.
At last, the difficulties of training the more difficult striker environments are discussed.

\subsection{On Improving the Sim-to-sim Transfer}


At first, the Unity simulation is still lacking the implementation of physical concepts like the car-to-car interaction and suffers from the reported (Section \ref{sec:alginment}) inaccuracies.
These can be further improved by putting more work into the simulation, but also other approaches are promising.
At the cost of more computational resources, domain randomization \cite{DomainRandomizationTobin2017} could achieve a more robust agent, potentially comprising an improved ability to generalize to the domain of Rocket League.
As the ground truth is provided by Rocket League, approaches from the field of supervised learning can be considered as well.

\subsection{Training under Human Conditions}


Once the physical domain gap is narrowed, the Unity simulation still does not consider training under human conditions.
Notably, the current observation space provides perfect information on the current state of the environment, whereas players in Rocket League have to cope with imperfect information due to solely perceiving the rendered image of the game.
Thus, the Unity simulation has to implement Rocket League's camera behavior as well.
However, one critical concern is that the RLBot API does not reveal the rendered image of Rocket League and therefore makes a transfer impossible as of now.
However, even if that information is made available by Psyonix, both simulations' visual appearances are very different.
The Unity simulation's aesthetics are very abstract, whereas Rocket League impresses with multiple arenas featuring many details concerning lighting, geometry, shaders, textures, particle effects, etc..
To overcome this gap of visual appearance, approaches of the previously described related work, like GraspGAN \cite{DBLP:conf/icra/BousmalisIWBKKD18}, can be considered.

Another challenge arises once the environment is partially observable.
It should be considered that the agent will probably need memory to be able to compete with human players.
Otherwise, the agent might not be able to capture the current affairs of its teammates and opponents.
For this purpose, multiple memory-based approaches might be suitable, like using a recurrent neural network or a transformer architecture.

Moreover, the multi-discrete action space used in this paper is a simplification of the original action space that features concurrent continuous and discrete actions.
Initially, the training was done using the PPO implementation of the ML-Agents toolkit \cite{Juliani2019}, which supports mixed (or hybrid) concurrent action spaces.
However, these experiments were quite unstable and hindered progress.
Therefore, Rocket League presents an interesting challenge for exploring such action spaces, of which other video games or applications are likely to take advantage.

\subsection{Difficulties of Training the harder Striker Environment}

While the goalie and the striker environment are relatively easy, the slightly more difficult striker one poses a much greater challenge due to multiple reasons:
\begin{itemize}
    \item Episodes are longer, leading to an even more delayed reward signal and more challenging credit assignment
    \item More states have to be explored by the agent
    \item Even more accurate touches on the ball have to be made to score
\end{itemize}
To overcome these problems, curriculum learning \cite{Bengio2009Currciculum} and reward shaping \cite{Gullapalli1992Shaping} can be considered.
In curriculum learning, the agent could face easier scenarios first and once success kicks in, the next level of difficulty can be trained.
However, catastrophic forgetting may occur and therefore a curriculum should sample from a distribution of scenarios to mitigate this issue.

Concerning reward shaping, multiple variants were casually tried without improving training results:
\begin{itemize}
    \item Reward the first touch on the ball
    \item Reward or penalize the distance between the ball and the agent
    \item Reward or penalize the dot product between the car's velocity and the direction from the car to the ball
\end{itemize}
Adding more reward signals along the agent's task introduces bias and is likely task-irrelevant.
For example, the agent could exploit such signals to cuddle with the ball at a close distance or to slowly approach the ball to maximize the cumulative return of the episode.
If those signals are turned off once the ball is touched, the value function might struggle to make further good estimates on the value of the current state of the environment, which ultimately may lead to misleading training experiences and hence an unstable learning process.
In spite of the results of these first explorative tests, future work shall examine whether these points shall be reconsidered.

\section{Conclusion}

Towards solving Rocket League by the means of Deep Reinforcement Learning, a fast simulation is crucial, because the original game cannot be sped up and neither parallelized on Linux-based clusters.
Therefore, we advanced the implementation of a Unity project that mimics the physical gameplay mechanics of Rocket League.
Although the implemented simulation is not perfectly accurate, we remarkably demonstrate that transferring a trained behavior from  Unity to Rocket League is robust and generalizes when dealing with a goalkeeper and striker task.
Hence, the sim-to-sim transfer is a suitable approach for learning agent behaviors in complex game environments.
After all, Rocket League still poses further challenges when targeting a complete match under human circumstances.
Based on our findings, we believe that Rocket League and its Unity counterpart will be valuable to various research fields and aspects, comprising: sim-to-sim transfer, partial observability, mixed action-spaces, curriculum learning, competitive and cooperative multi-agent settings.

\bibliographystyle{IEEEtran}
\bibliography{IEEEabrv, bibliography/bibliography.bib, bibliography/related_work.bib, bibliography/fundamentals.bib}
\end{document}